# Rapid Adaptation of POS Tagging for Domain Specific Uses

John E. Miller[1]     Michael Bloodgood[1]     Manabu Torii[2]     K. Vijay-Shanker[1]

[1]Computer & Information Sciences
University of Delaware
Newark, DE 19716
{jmiller,bloodgoo,vijay}@`cis.udel.edu`

[2]Biostatistics, Bioinformatics and Biomathematics
Georgetown University Medical Center
Washington, DC 20057
`mt352@georgetown.edu`

## 1 Introduction

Part-of-speech (POS) tagging is a fundamental component for performing natural language tasks such as parsing, information extraction, and question answering. When POS taggers are trained in one domain and applied in significantly different domains, their performance can degrade dramatically. We present a methodology for rapid adaptation of POS taggers to new domains. Our technique is unsupervised in that a manually annotated corpus for the new domain is not necessary. We use suffix information gathered from large amounts of raw text as well as orthographic information to increase the lexical coverage. We present an experiment in the Biological domain where our POS tagger achieves results comparable to POS taggers specifically trained to this domain.

Many machine-learning and statistical techniques employed for POS tagging train a model on an annotated corpus, such as the Penn Treebank (Marcus et al, 1993). Most state-of-the-art POS taggers use two main sources of information: 1) Information about neighboring tags, and 2) Information about the word itself. Methods using both sources of information for tagging are: Hidden Markov Modeling, Maximum Entropy modeling, and Transformation Based Learning (Brill, 1995).

In moving to a new domain, performance can degrade dramatically because of the increase in the unknown word rate as well as domain-specific word use. We improve tagging performance by attacking these problems. Since our goal is to employ minimal manual effort or domain-specific knowledge, we consider only orthographic, inflectional and derivational information in deriving POS. We bypass the time, cost, resource, and content expert intensive approach of annotating a corpus for a new domain.

## 2 Methodology and Experiment

The initial components in our POS tagging process are a lexicon and part of speech (POS) tagger trained on a generic domain corpus. The lexicon is updated to include domain specific information based on suffix rules applied to an un-annotated corpus. Documents in the new domain are POS tagged using the updated lexicon and orthographic information. So, the POS tagger uses the domain specific updated lexicon, along with what it knows from generic training, to process domain specific text and output POS tags.

In demonstrating feasibility of the approach, we used the fnTBL-1.0 POS tagger (Ngai and Florian, 2001) based on Brill's Transformation Based Learning (Brill, 1995) along with its lexicon and contextual rules trained on the Wall Street Journal corpus.

To update the lexicon, we processed 104,322 abstracts from five of the 500 compressed data files in the 2005 PubMed/Medline database (Smith *et a*l, 2004). As a result of this update, coverage of words with POS tags from the lexicon increased from 73.0% to 89.6% in our test corpus.

Suffix rules were composed based on information from Michigan State University's Suffixes and Parts of Speech web page for Graduate Record Exams (DeForest, 2000). The suffix endings indicate the POS used for new words. However, as seen in the table of suffix examples below, there can be significant lack of precision in assigning POS based just on suffixes.

| Suffix | POS | #uses/ %acc |
|---|---|---|
| ize; izes | VB VBP; VBZ | 23/100% |
| ous | JJ | 195/100% |
| er, or; ers, ors | NN; NNS | 1471/99.5% |
| ate; ates | VB VBP | 576/55.7% |

Most suffixes did well in determining the actual POS assigned to the word. Some such as "-er" and "-or" had very broad use as well. "-ate" typically forms a verb from a noun or adjective in a generic domain. However in scientific domains it often indicates a noun or adjective word form. (In work just begun, we add POS assignment confirmation tests to suffix rules so as to confirm POS tags while maintaining our domain independent and unsupervised analysis of un-annotated corpora.)

Since the fnTBL POS tagger gives preliminary assignment of POS tags based on the first POS listed for that word in the lexicon, it is vital that the first POS tag for a common word be correct. Words ending in '-ing' can be used in a verbal (VBG), adjectival (JJ) or noun (NN) sense. Our intuition is that the '-ed' form should also appear often when the verbal sense dominates. In contrast, if the ratio heavily favors the '-ing' form then we expect the noun sense to dominate.

We incorporated this reasoning into a computationally defined process which assigned the NN tag first to the following words: *binding, imaging, learning, nursing, processing, screening, signaling, smoking, training, and underlying*. Only *underlying* seems out of place in this list.

In addition to inflectional and derivational suffixes, we used rules based on orthographic characteristics. These rules defined proper noun and number or code categories.

## 3  Results and Conclusion

For testing purposes, we used approximately half the abstracts of the GENIA corpus (version 3.02) described in (Tateisi *et al*, 2003). As the GENIA corpus does not distinguish between common and proper nouns we dropped that distinction in evaluating tagger performance.

POS tagging accuracy on our GENIA test set (second half of abstracts) consisting of 243,577 words is shown in the table below.

| Source | Accuracy |
|---|---|
| Original fnTBL lexicon | 92.58% |
| Adapted lexicon (Rapid) | 94.13% |
| MedPost | 94.04% |
| PennBioIE[1] | 93.98% |

---

[1] Note that output from the tagger is not fully compatible with GENIA annotation.

The original fnTBL tagger has an accuracy of 92.58% on the GENIA test corpus showing that it deals well with unknown words from this domain. Our rapid adaptation tagger achieves a modest 1.55% absolute improvement in accuracy, which equates to a 21% error reduction.

There is little difference in performance between our rapid adaptation tagger and the MedPost (Smith *et al*, 2004) and PennBioIE (Kulick *et al*, 2004) taggers. The PennBioIE tagger employs maximum entropy modeling and was developed using 315 manually annotated Medline abstracts. The MedPost tagger also used domain-specific annotated corpora and a 10,000 word lexicon, manually updated with POS tags.

We have improved the accuracy of the fnTBL-1.0 tagger for a new domain by adding words and POS tags to its lexicon via unsupervised methods of processing raw text from the new domain. The accuracy of the resulting tagger compares well to those that have been trained to this domain using annotation effort and domain-specific knowledge.